# Normalized Web Distance and Word Similarity

Rudi L. Cilibrasi and Paul M.B. Vitányi *

September 5, 2018

## 1 Introduction

Objects can be given literally, like the literal four-letter genome of a mouse, or the literal text of *War and Peace* by Tolstoy. For simplicity we take it that all meaning of the object is represented by the literal object itself. Objects can also be given by name, like "the four-letter genome of a mouse," or "the text of *War and Peace* by Tolstoy." There are also objects that cannot be given literally, but only by name, and that acquire their meaning from their contexts in background common knowledge in humankind, like "home" or "red."

To make computers more intelligent one would like to represent meaning in computer-digestible form. Long-term and labor-intensive efforts like the *Cyc* project [24] and the *WordNet* project [17] try to establish semantic relations between common objects, or, more precisely, *names* for those objects. The idea is to create a semantic web of such vast proportions that rudimentary intelligence, and knowledge about the real world, spontaneously emerge. This comes at the great cost of designing structures capable of manipulating knowledge, and entering high quality contents in these structures by knowledgeable human experts. While the efforts are long running and large scale, the overall information entered is minute compared to what is available on the Internet.

The rise of the Internet has enticed millions of users to type in trillions of characters to create billions of web pages of on average low quality contents. The sheer mass of the information about almost every conceivable topic makes it likely that extremes will cancel and the majority or average is meaningful in a low-quality approximate sense.

The goal of this chapter is to introduce the normalized web distance (NWD) method to determine similarity between words and phrases. It is a general way to tap the amorphous low-grade knowledge available for free on the Internet, typed in by local users aiming at personal gratification of diverse objectives, and yet globally achieving what is effectively the largest semantic electronic database in the world. Moreover, this database is available for all by using any search engine that can return aggregate page-count estimates for a large range of search-queries. In the paper [10] introducing the NWD it was called 'normalized Google distance (NGD),' but since Google doesn't allow computer searches anymore, we opt for the more neutral and descriptive NWD.

Previously, a compression-based method was developed to establish a universal similarity metric among objects given as finite binary strings [3, 27, 28, 11, 9, 21, 34], which was widely reported [31, 32, 15] and has led to hundreds of applications in research as reported by Google Scholar. The objects can be genomes [27, 28, 10], music pieces in MIDI format [11, 10], computer programs in Ruby or C, pictures in simple bitmap formats, astronomical data, literature, [10], time sequences such as heart rhythm data [21, 34], and so on. The method is feature free in the sense that it doesn't analyze the files looking for particular features; rather it analyzes all features simultaneously and determines the similarity between every pair of objects according to the most dominant shared feature. It is not parameter laden, in fact, there are no parameters to set. In the genomic context it is alignment free and much faster than alignment methods; it provides an alignment-free method such as looked for in many genomic problems.

*Rudi Cilibrasi was supported in part by the Netherlands BSIK/BRICKS project, and by NWO project 612.55.002. Address: CWI, Science Park 123, 1098 XG Amsterdam, The Netherlands. Email: cilibrar@gmail.com. Supported in part by the EU NoE PASCAL, and the Netherlands BSIK/BRICKS project. Address: CWI, Science Park 123, 1098 XG Amsterdam, The Netherlands. Email: Paul.Vitanyi@cwi.nl.

But in the case of word similarity we do not have the objects themselves. Rather, we have names for objects, or other words, and the crucial point is that the compression method described above analyzes the objects themselves. This precludes comparison of abstract notions or other objects that don't lend themselves to direct analysis, like emotions, colors, Socrates, Plato, Mike Bonanno and Albert Einstein. While the method that compares the objects themselves is particularly suited to obtain knowledge about the similarity of objects themselves, irrespective of common beliefs about such similarities, the normalized web distance method of [10] uses only the name of an object (or even more simply words and phrases), and obtains knowledge about the similarity of objects (or the words and phrases), by tapping available information generated by multitudes of web users.

In [10] the following example experiment determining word similarity by the normalized web distance method is described. At that time, a google search for "horse", returned 46,700,000 hits. The number of hits for the search term "rider" was 12,200,000. Searching for the pages where both "horse" and "rider" occur gave 2,630,000 hits, and Google indexed 8,058,044,651 web pages at the time. Using these numbers in the main formula (8) we derive below, with $N = 8,058,044,651$, this yields a normalized web distance, denoted by $e_G(\cdot,\cdot)$, between the terms "horse" and "rider" as follows:

$$e_G(\text{horse}, \text{rider}) \approx 0.443.$$

We did the same calculation when Google indexed only half the number of pages: 4,285,199,774. It is instructive that the probabilities of the used search terms didn't change significantly in the meantime: with half the number of pages indexed, the number of hits for "horse" was 23,700,000, for "rider" it was 6,270,000, and for "horse, rider" it was 1,180,000. The $e_G(\text{horse}, \text{rider})$ we computed in that situation was $\approx 0.460$. This is in line with our contention that the relative frequencies of web pages containing search terms gives objective information about the semantic relations between the search terms.

# 2 Some Methods for Word Similarity

There is a great deal of work in cognitive psychology [25], linguistics, and computer science, about using word (or phrase) frequencies in context in text corpora to develop measures for word similarity or word association, partially surveyed in [36, 35], going back to at least the 1960s [26]. Some issues in word similarity are association measures, attributed word similarity, and relational word similarity.

## 2.1 Association Measures

Association measures were the subject of [37, 38]. There the algorithm used is called PMI-IR, short for pointwise mutual information (PMI), to analyze statistical data collected by information retrieval (IR). The PMI-IR algorithm in [37], like LSA discussed in Section 2.4, and in fact the NWD method which forms the core of this chapter, is based on co-occurrence of words. Assume we are given name1 and we are looking which name2 is closest related. Essentially the algorithm uses the idea that the relatedness of name2 to name1 is expressed by

$$\text{score}(\text{name2}) = \log \frac{\Pr(\text{name2} \ \& \ \text{name1})}{\Pr(\text{name1}) \Pr(\text{name2})}.$$

Here the precise meaning of the connective "&" is subject to refinements as below. It can mean "in the same page" or "occurrences near one another in a certain window size," and so on. Since the method is looking for the maximum score one can drop the logarithm and Pr(name1) (because name1 is fixed). Thus, the formula simplifies to looking for name2 that maximizes

$$\frac{\Pr(\text{name2} \ \& \ \text{name1})}{\Pr(\text{name2})}. \tag{1}$$

This leaves the question of how to compute the probabilities. This is done using a search engine and the Internet. The search engine used in the reference is Altavista, and four different probabilities are presented. Note that because we are looking at a ratio we need only the number of hits of Altavista for a given search term.

- In the simplest case we consider co-occurrence when the two words occur in the same page:

$$\text{score}_1(\text{name2}) = \frac{\text{hits(name1 AND name2)}}{\text{hits(name2)}}.$$

- The next method asks if the words occur near each other:

$$\text{score}_2(\text{name2}) = \frac{\text{hits(name1 NEAR name2)}}{\text{hits(name2)}}.$$

Two other methods in increasing degree of sophistication are presented to refine the hits with AND, NOT, NEAR in both numerator and denominator.

Related preceding work in [4] defines a notion of 'interest' to determine interesting associations in large databases. The *interest* of the association between $A$ and $B$ is defined exactly like (1) without the logarithm. This has apparently been used for data mining but not for text mining.

We continue with the results in [37]. Experiments were done on synonym test questions from the Test of English as a Foreign Language (TOEFL) and 50 synonym test questions from a collection of tests for students of English as a Second Language (ESL). On both tests, the algorithm obtains a score of 74%. PMI-IR is contrasted with Latent Semantic Analysis (LSA), Section 2.4, which achieves a score of 64% on the same 80 TOEFL questions. Reference [37] notes that PMI-IR uses a much larger data source than LSA and PMI-IR (in all of the scores except for $\text{score}_1$) uses a much smaller chunk (text window examined) size than LSA. A similar application using the same method is used in [38] with some more or less evident refinements to classify 410 reviews from the website Epinions sampled from four different domains (reviews of automobiles, banks, movies, and travel destinations). The accuracy ranges from 84% for automobile reviews to 66% for movie reviews.

## 2.2 Attributes

Here the approach is to determine attributes as representation of words. Consider a target noun, say "horse," and this noun occurs in a sentence as "the rider rides the horse". Then we have the triple $(\text{rider}, \text{rides}, \text{horse})$ and the pair $(\text{rider}, \text{rides})$ is an attribute of the noun "horse." This approach is then coupled with an appropriate word similarity measure like the one discussed based on pointwise mutual information, or another one like the cosine similarity measure in LSA as in Section 2.4. In fact, LSA is an example of attributional similarity of words. Good references are [30, 16, 18].

## 2.3 Relational Word Similarity

We cite [39]: "Relational similarity is correspondence between relations, in contrast with attributional similarity, which is correspondence between attributes. When two words have a high degree of attributional similarity, we call them synonyms. When two pairs of words have a high degree of relational similarity, we say that their relations are analogous. For example, the word pair mason:stone is analogous to the pair carpenter:wood." In this context, LSA as in Section 2.4 measures similarity between two words but not between two relations between two pairs of words. One way to measure similarity between the relatedness of pairs of words is to score the relation between a pair of words as frequencies of features (predefined patterns in a large corpus) in vectors and then compare the closeness of the respective vectors by measuring the distance according to the Euclidean distance, the cosine between the vectors, or the logarithm of the cosine. Often a search engine like Altavista or Google is used to determine the frequency information to build the vectors. In [39] the author introduces a new method 'latent relational analysis (LRA)' that uses a search engine, a thesaurus of synonyms, and single value decomposition or SVD. For SVD see the discussion on LSA below. LRA is an involved method, and for more details we refer the reader to the cited reference.

## 2.4 Latent Semantic Analysis

Most of the approaches have tackled synonymy detection based on the vector space model and/or probabilistic models. Obviously, there exist many other works for other semantic relations. One of the most successful is Latent Semantic Analysis (LSA) [25] that has been applied in various forms in a great number of applications. The basic assumption of Latent Semantic Analysis is that "the cognitive similarity between any two words is reflected in the way they co-occur in small subsamples of the language." In particular, this is implemented by constructing a matrix with rows labeled by the $d$ documents involved, and the columns labeled by the $a$ attributes (words, phrases). The entries are the number of times the column attribute occurs in the row document. The entries are then processed by taking the logarithm of the entry and dividing it by the number of documents the attribute occurred in, or some other normalizing function. This results in a sparse but high-dimensional matrix $A$. A main feature of LSA is to reduce the dimensionality of the matrix by projecting it into an adequate subspace of lower dimension using singular value decomposition $A = UDV^T$ where $U, V$ are orthogonal matrices and $D$ is a diagonal matrix. The diagonal elements $\lambda_1, \ldots, \lambda_p$ ($p = \min\{d, a\}$) satisfy $\lambda_1 \geq \cdots \geq \lambda_p$, and the closest matrix $A_k$ of dimension $k < \text{Rank}(A)$ in terms of the so-called Frobenius norm is obtained by setting $\lambda_i = 0$ for $i > k$. Using $A_k$ corresponds to using the most important dimensions. Each attribute is now taken to correspond to a column vector in $A_k$, and the similarity between two attributes is usually taken to be the cosine between their two vectors.

To compare LSA to the method of using the normalized web distance (NWD) of [10] we treat in detail below, the documents could be the web pages, the entries in matrix $A$ are the frequencies of a search term in each web page. This is then converted as above to obtain vectors for each search term. Subsequently, the cosine between vectors gives the similarity between the terms. LSA has been used in a plethora of applications ranging from database query systems to synonymy answering systems in TOEFL tests. Comparing LSA's performance to the NWD performance is problematic for several reasons. First, the numerical quantity measuring the semantic distance between pairs of terms cannot directly be compared, since they have quite different epistimologies. Indirect comparison could be given using the method as basis for a particular application, and comparing accuracies. However, application of LSA in terms of the web using a search engine is computationally out of the question, because the matrix $A$ would have $10^{10}$ rows, even if the search engine would report frequencies of occurrences in web pages and identify the web pages properly. One would need to retrieve the entire web database, which is many terabytes. Moreover, each invocation of a web search takes a significant amount of time, and we cannot automatically make more than a certain number of them per day. An alternative interpretation by considering the web as a single document makes the matrix $A$ above into a vector and appears to defeat the LSA process altogether. Summarizing, the basic idea of our method is similar to that of LSA in spirit. What is novel is that we can do it with selected terms over a very large document collection, whereas LSA involves matrix operations over a closed collection of limited size, and hence is not possible to apply in the web context.

As with LSA, many other previous approaches of extracting correlations from text documents are based on text corpora that are many orders of magnitudes smaller, and that are in local storage, and on assumptions that are more refined, than what we propose here. In contrast, [37, 38, 12, 2] and the many references cited there, use the web and search engine page counts to identify lexico-syntactic patterns or other data. Again, the theory, aim, feature analysis, and execution are different from ours.

# 3 Background of the NWD Method

The NWD method below automatically extracts semantic relations between arbitrary objects from the web in a manner that is feature free, up to the search engine used, and computationally feasible. This is a new direction of feature-free and parameter-free data mining. Since the method is parameter-free it is versatile and as a consequence domain, genre, and language independent.

The main thrust in [10] is to develop a new theory of semantic distance between a pair of objects, based on (and unavoidably biased by) a background contents consisting of a database of documents. An example of the latter is the set of pages constituting the Internet. Another example would be the set of all ten-word phrases generated by a sliding window passing over the text in a database of web pages.

Similarity relations between pairs of objects are distilled from the documents by just using the number of documents in which the objects occur, singly and jointly. These counts may be taken with regard to location, that is, we consider a sequence of words, or without regard to location which means we use a bag of words. They may be taken with regard to multiplicity in a term frequency vector or without regard to multiplicity in a binary term vector, as the setting dictates. These decisions determine the normalization factors and feature classes that are analyzed, but do not alter substantially the structure of the algorithm. For us, the semantics of a word or phrase consists of the set of web pages returned by the query concerned. Note that this can mean that terms with different meanings have the same semantics, and that opposites like "true" and "false" often have a similar semantics. Thus, we just discover associations between terms, suggesting a likely relationship.

As the web grows, the semantics may become less primitive. The theoretical underpinning is based on the theory of Kolmogorov complexity [29], and is in terms of coding and compression. This allows to express and prove properties of absolute relations between objects that cannot be expressed by other approaches. We start with a technical introduction outlining some notions underpinning our approach: Kolmogorov complexity (Section 4), and information distance resulting in the compression-based similarity metric (Section 5). In Section 6 we give the theoretic underpinning of the normalized web distance. In Section 7.1 and Section 7.2 we present clustering and classification experiments to validate the universality, robustness, and accuracy of the normalized web distance. In Section 7.3 we present a toy example of translation. In Section 7.4 we test repetitive automatic performance of the normalized web distance against uncontroversial semantic knowledge: We present the results of a massive randomized classification trial we conducted to gauge the accuracy of our method against the expert knowledge implemented over decades in the WordNet database. The preliminary publication [10] of this work in the web archive ArXiv was widely reported and discussed, for example [19, 13]. The actual experimental data can be downloaded from [8]. The method is implemented as an easy-to-use software tool [7], free for commercial and non-commercial use according to a BSD style license.

The application of the theory we develop is a method that is justified by the vastness of the Internet, the assumption that the mass of information is so diverse that the frequencies of pages returned by a good set of search engine queries averages the semantic information in such a way that one can distill a valid semantic distance between the query subjects. The method starts from scratch, is feature-free in that it uses just the web and a search engine to supply contents, and automatically generates relative semantics between words and phrases. As noted in [2], the returned counts can be inaccurate although linguists judge the accuracy of for example Google counts trustworthy enough [35]. In [20] (see also the many references to related research) it is shown that web searches for rare two-word phrases correlated well with the frequency found in traditional corpora, as well as with human judgments of whether those phrases were natural. Thus, search engines on the web are the simplest means to get the most information. The experimental evidence provided here shows that our method yields reasonable results, gauged against common sense ('colors' are different from 'numbers') and against the expert knowledge in the WordNet database.

# 4 Brief Introduction to Kolmogorov Complexity

The basis of much of the theory explored in this paper is Kolmogorov complexity [22]. For an introduction and details see the textbook [29]. Here we give some intuition and notation. We assume a fixed reference universal programming system. Such a system may be a general computer language like LISP or Ruby, and it may also be a fixed reference universal Turing machine $U$ in a given standard enumeration of Turing machines $T_1, T_2, \ldots$ of the type such that $U(i,p) = T_i(p) < \infty$ for every index $i$ and program $p$. This also involves that $U$ started on input $(i,p)$ and $T_i$ started on input $p$ both halt after a finite number of steps, which may be different in both cases and possibly unknown. Such $U$'s have been called 'optimal' [22]. The last choice has the advantage of being formally simple and hence easy to theoretically manipulate. But the choice makes no difference in principle, and the theory is invariant under changes among the universal programming systems, provided we stick to a particular choice. We only consider programs that are binary finite strings and such that for every Turing machine the set of programs is a prefix-free set or *prefix code*: no program is a proper prefix of another program for this Turing machine. Thus, universal programming systems are such that the associated set of programs is a prefix code—as is the case in all standard computer languages.

The *Kolmogorov complexity* $K(x)$ of a string $x$ is the length, in bits, of a shortest computer program (there may be more than one) of the fixed reference computing system, such as a fixed optimal universal Turing machine that (without input) produces $x$ as output. The choice of computing system changes the value of $K(x)$ by at most an additive fixed constant. Since $K(x)$ goes to infinity with $x$, this additive fixed constant is an ignorable quantity if we consider large $x$. Given the fixed reference computing system, the function $K$ is not computable.

One way to think about the Kolmogorov complexity $K(x)$ is to view it as the length, in bits, of the ultimate compressed version from which $x$ can be recovered by a general decompression program. Compressing $x$ using the compressor *gzip* results in a file $x_g$ with (for files that contain redundancies) the length $|x_g| < |x|$. Using a better compressor *bzip2* results in a file $x_b$ with (for redundant files) usually $|x_b| < |x_g|$; using a still better compressor like *PPMZ* results in a file $x_p$ with (for again appropriately redundant files) $|x_p| < |x_b|$. The Kolmogorov complexity $K(x)$ gives a lower bound on the ultimate length of a compressed version for every existing compressor, or compressors that are possible but not yet known: the value $K(x)$ is less or equal to the length of every effectively compressed version of $x$. That is, $K(x)$ gives us the ultimate value of the length of a compressed version of $x$ (more precisely, from which version $x$ can be reconstructed by a general purpose decompresser), and our task in designing better and better compressors is to approach this lower bound as closely as possible.

Similarly, we can define the *conditional Kolmogorov complexity* $K(x|y)$ as the length of a shortest program that computes output $x$ given input $y$, and the *joint Kolmogorov complexity* $K(x, y)$ as the length of a shortest program that without input computes the pair $x, y$ and a way to tell them apart.

Definition 1 A *computable* rational valued function is one that can be computed by a halting program on the reference universal Turing machine. A function $f$ with real values is *upper semicomputable* if there is a computable rational valued function $\phi(x, k)$ such that $\phi(x, k+1) \leq \phi(x, k)$ and $\lim_{k\to\infty} \phi(x, k) = f(x)$; it is *lower semicomputable* if $-f$ is upper semicomputable. We call a real valued function $f$ *computable* if it is both lower semicomputable and upper semicomputable.

It has been proved [29] that the Kolmogorov complexity is the least upper semicomputable code length up to an additive constant term. Clearly, every Turing machine $T_i$ defines an upper semicomputable code length of a source word $x$ by $\min_q\{|q| : T_i(q) = x\}$. For every $i$ there is a constant $c_i$ such that for every $x$ we have $\min_p\{|p| : U(p) = x\} \leq \min_q\{|q| : T_i(q) = x\} + c_i$.

An important identity is the *symmetry of information*

$$K(x, y) = K(x) + K(y|x) = K(y) + K(x|y), \tag{2}$$

which holds up to an $O(\log K(xy))$ additive term.

The following notion is crucial in the later sections. We define the *universal probability* $\mathbf{m}$ by

$$\mathbf{m}(x) = 2^{-K(x)}, \tag{3}$$

which satisfies $\sum_x \mathbf{m}(x) \leq 1$ by the Kraft inequality [23, 14, 29] since $\{K(x) : x \in \{0,1\}^*\}$ is the length set of a prefix code. To obtain a proper probability mass function we can concentrate the surplus probability on a new undefined element $u$ so that $\mathbf{m}(u) = 1 - \sum_x \mathbf{m}(x)$. The universal probability mass function $\mathbf{m}$ is a form of *Occam's razor* since $\mathbf{m}(x)$ is high for simple objects $x$ whose $K(x)$ is low such as $K(x) = O(\log |x|)$, and $\mathbf{m}(y)$ is low for complex objects $y$ whose $K(y)$ is high such as $K(y) \geq |y|$.

It has been proven [29] that $\mathbf{m}$ is the greatest lower semicomputable probability mass function up to a constant multiplicative factor. Namely, it is easy to see that $\mathbf{m}$ is a lower semicomputable probability mass function, and it turns out that for every lower semicomputable probability mass function $P$ there is a constant $c_P$ such that for every $x$ we have $c_P\mathbf{m}(x) \geq P(x)$.

# 5 Information Distance

In [3] the following notion is considered: given two strings $x$ and $y$, what is the length of the shortest binary program in the reference universal computing system such that the program computes output $y$ from input $x$, and also output $x$ from input $y$. This is called the *information distance*. It turns out that, up to a negligible logarithmic additive term, it is equal to

$$E(x, y) = \max\{K(x|y), K(y|x)\}. \tag{4}$$

We now discuss the important properties of $E$.

Definition 2 A distance $D(x,y)$ is a *metric* if $D(x,x) = 0$ and $D(x,y) > 0$ for $x \neq y$; $D(x,y) = D(y,x)$ (symmetry); and $D(x,y) \leq D(x,z) + D(z,y)$, (triangle inequality) for all $x, y, z$.

For a distance function or metric to be reasonable, it has to satisfy an additional condition, referred to as *density condition*. Intuitively this means that for every object $x$ and positive real value $d$ there is at most a certain finite number of objects $y$ at distance $d$ from $x$. This requirement excludes degenerate distance measures like $D(x,y) = 1$ for all $x \neq y$. Exactly how fast we want the distances of the strings $y$ from $x$ to go to infinity is not important, it is only a matter of scaling. For convenience, we will require the following *density conditions*:

$$\sum_{y:y\neq x} 2^{-D(x,y)} \leq 1 \text{ and } \sum_{x:x\neq y} 2^{-D(x,y)} \leq 1 \ . \tag{5}$$

Finally, we allow only distance measures that are computable in some broad sense, which will not be seen as unduly restrictive. The upper semicomputability in Definition 1 is readily extended to two-argument functions and in the present context to distances. We require the distances we deal with to be upper semicomputable. This is reasonable: if we have more and more time to process $x$ and $y$, then we may discover newer and newer similarities among them, and thus may revise our upper bound on their distance.

Definition 3 An *admissible distance* is a total, possibly asymmetric, nonnegative function with real values on the pairs $x, y$ of binary strings that is 0 if and only if $x = y$, is upper semicomputable, and satisfies the density requirement (5).

Definition 4 Consider a family $\mathcal{F}$ of two-argument real valued functions. A function $f$ is *universal* for the family $\mathcal{F}$ if for every $g \in \mathcal{F}$ we have

$$f(x,y) \leq g(x,y) + c_g,$$

where $c_g$ is a constant that depends only on $g$ but not on $x, y$ and $f$. We say that $f$ *minorizes* every $g \in \mathcal{F}$ up to an additive constant.

The following theorem is proven in [3, 29].

Theorem 1 (i) *$E$ is universal for the family of admissible distances.*
(ii) *$E$ satisfies the metric (in)equalities up to an $O(1)$ additive term.*

If two strings $x$ and $y$ are close according to *some* admissible distance $D$, then they are at least as close according to the metric $E$. Every feature in which we can compare two strings can be quantified in terms of a distance, and every distance can be viewed as expressing a quantification of how much of a particular feature the strings do not have in common (the feature being quantified by that distance). Therefore, the information distance is an admissible distance between two strings minorizing the *dominant* feature expressible as an admissible distance which they have in common. This means that, if we consider more than two strings, the information distance between every pair may be based on minorizing a different dominant feature.

## 5.1 Normalized Information Distance

If strings of length $1,000$ bits differ by 800 bits then these strings are very different. However, if two strings of $1,000,000$ bits differ by 800 bits only, then they are very similar. Therefore, the information distance itself is not suitable to express true similarity. For that we must define a relative information distance: we need to normalize the information distance. Our objective is to normalize the universal information distance $E$ in (4) to obtain a universal *similarity* distance. It should give a similarity with distance 0 when objects are maximally similar and distance 1 when they are maximally dissimilar. Such an approach was first proposed in [27] in the context of genomics-based phylogeny, and improved in [28] to the one we use here. Several alternative ways of normalizing the information distance do not work. It is paramount that the normalized version of the information metric is also a metric in the case we deal with literal objects that contain all their properties within. Were it not, then the relative relations between the objects would be disrupted and this could lead to anomalies, if, for instance, the triangle inequality would be violated for the normalized version. However, for nonliteral objects that have a semantic distance NWD based on hit count statistics as in Section 6, which is the real substance of this work, the triangle inequality will be seen not to hold.

The *normalized information distance* (NID) is defined by

$$e(x,y) = \frac{\max\{K(x|y), K(y|x)\}}{\max\{K(x), K(y)\}}. \tag{6}$$

THEOREM 2 *The normalized information distance* $e(x,y)$ *takes values in the range* $[0,1]$ *and is a metric, up to ignorable discrepancies.*

The theorem is proven in [28] and the ignorable discrepancies are additive terms $O((\log K)/K)$ where $K$ is the maximum of the Kolmogorov complexities of strings $x, y, z$ involved in the metric (in)equalities. The NID discovers for every pair of strings the feature in which they are most similar, and expresses that similarity on a scale from 0 to 1 (0 being the same and 1 being completely different in the sense of sharing no features). It has several wonderful properties that justify its description as the most informative metric [28].

### 5.2 Normalized Compression Distance

The normalized information distance $e(x,y)$, which we call 'the' similarity metric because it accounts for the dominant similarity between two objects, is not computable since the Kolmogorov complexity is not computable. First we observe that using $K(x,y) = K(xy) + O(\log \min\{K(x), K(y)\}$ and the symmetry of information (2) we obtain

$$\max\{K(x|y), K(y|x)\} = K(xy) - \min\{K(x), K(y)\},$$

up to an additive logarithmic term $O(\log K(xy))$, which we ignore in the sequel. In order to use the NID in practice, admittedly with a leap of faith, the approximation of the Kolmogorov complexity uses real compressors to approximate the Kolmogorov complexities $K(x), K(y), K(xy)$. A compression algorithm defines a computable function from strings to the lengths of the compressed versions of those strings. Therefore, the number of bits of the compressed version of a string is an upper bound on Kolmogorov complexity of that string, up to an additive constant depending on the compressor but not on the string in question. This direction has yielded a very practical success of Kolmogorov complexity. Substitute the last displayed equation in the NID of (6), and subsequently use a real-world compressor Z (such as *gzip, bzip2*, and *PPMZ*) to heuristically replace the Kolmogorov complexity. In this way, we obtain the distance $e_Z$, often called the *normalized compression distance* (NCD), defined by

$$e_Z(x,y) = \frac{Z(xy) - \min\{Z(x), Z(y)\}}{\max\{Z(x), Z(y)\}}, \tag{7}$$

where $Z(x)$ denotes the binary length of the compressed version of the file $x$, compressed with compressor Z. The distance $e_Z$ is actually a family of distances parametrized with the compressor Z. The better Z is, the closer $e_Z$ approaches the normalized information distance, the better the results. Since Z is computable the distance $e_Z$ is computable. In [9] it is shown that under mild conditions on the compressor Z, the distance $e_Z$ takes values in $[0,1]$ and is a metric, up to negligible errors. One may imagine $e$ as the limiting case $e_K$, where $K(x)$ denotes the number of bits in the shortest code for $x$ from which $x$ can be decompressed by a computable general purpose decompressor.

## 6 Word Similarity: Normalized Web Distance

Can we find an equivalent of the normalized information distance for names and abstract concepts? In [10] the formula (6) to determine word similarity from the Internet is derived. It is also proven that the distance involved is 'universal' in a precise quantified manner. The present approach follows the treatment in [29] and obtains 'universality' in yet another manner by viewing the normalized web distance (8) below as a computable approximation to the universal distribution $\mathbf{m}$ of (3).

Let $\mathbf{W}$ be the set of pages of the Internet, and let $\mathbf{x} \subseteq \mathbf{W}$ be the set of pages containing the search term $x$. By the conditional version of (3) in [29], which appears straightforward but is cumbersome to explain here, we have $\log 1/\mathbf{m}(\mathbf{x}|\mathbf{x} \subseteq \mathbf{W}) = K(\mathbf{x}|\mathbf{x} \subseteq \mathbf{W}) + O(1)$. This equality relates the incompressibility of the set of pages on the web containing a given search term to its universal probability. We know that $\mathbf{m}$ is lower semicomputable since $K$ is upper semicomputable, and $\mathbf{m}$ is not computable since $K$ is not computable. While we cannot compute $\mathbf{m}$, a natural heuristic is to use the distribution of $x$ on the web to approximate $\mathbf{m}(\mathbf{x}|\mathbf{x} \subseteq \mathbf{W})$. Let us define the probability mass function $g(x)$ to be the probability that the search term $x$ appears in a page indexed by a given Internet search engine G, that is, the number of pages returned divided by the number $N$ which is the sum of the numbers of occurrences of search terms in each page, summed over all pages indexed. Then the Shannon–Fano code [29] length associated with $g$ can be set at

$$G(x) = \log \frac{1}{g(x)}.$$

Replacing $Z(x)$ by $G(x)$ in the formula in (7), we obtain the distance $e_G$, called the *normalized web distance* (NWD), which we can view as yet another approximation of the normalized information distance, defined by

$$\begin{aligned} e_G(x,y) &= \frac{G(xy) - \min\{G(x), G(y)\}}{\max\{G(x), G(y)\}} \qquad (8) \\ &= \frac{\max\{\log f(x), \log f(y)\} - \log f(x,y)}{\log N - \min\{\log f(x), \log f(y)\}}, \end{aligned}$$

where $f(x)$ is the number of pages containing $x$, the frequency $f(x,y)$ is the number of pages containing both $x$ and $y$, and $N$ is defined above.

Since the code $G$ is a Shannon-Fano code for the probability mass function $g$ it yields an on average minimal code-word length. This is not so good as an individually minimal code-word length, but is an approximation to it. Therefore, we can view the search engine G as a compressor using the web, and $G(x)$ as the binary length of the compressed version of the set of all pages containing the search term $x$, given the indexed pages on the web. The distance $e_G$ is actually a family of distances parametrized with the search engine G.

The better a search engine G is in the sense of covering more of the Internet and returning more accurate aggregate page counts, the closer $e_G$ approaches the normalized information distance $e$ of (6), with $K(x)$ replaced by $K(\mathbf{x}|\mathbf{x} \subseteq \mathbf{W})$ and similarly the other terms, and the better the results are expected to be.

In practice, we use the page counts returned by the search engine for the frequencies and choose $N$. From (8) it is apparent that by increasing $N$ we decrease the NWD, everything gets closer together, and by decreasing $N$ we increase the NWD, everything gets further apart. Our experiments suggest that every reasonable value can be used as normalizing factor $N$, and our results seem in general insensitive to this choice. This parameter $N$ can be adjusted as appropriate, and one can often use the number of indexed pages for $N$. $N$ may be automatically scaled and defined as an arbitrary weighted sum of common search term page counts.

The better G is the more informative the results are expected to be. In [10] it is shown that the distance $e_G$ is computable and is symmetric, that is, $e_G(x,y) = e_G(y,x)$. It only satisfies “half” of the identity property, namely $e_G(x,x) = 0$ for all $x$, but $e_G(x,y) = 0$ can hold even if $x \neq y$, for example, if the terms $x$ and $y$ always occur together in a web page.

The NWD also does *not* satisfy the triangle inequality $e_G(x,y) \leq e_G(x,z) + e_G(z,y)$ for all $x, y, z$. To see that, choose $x$, $y$, and $z$ such that $x$ and $y$ never occur together, $z$ occurs exactly on those pages on which $x$ or $y$ occurs, and $f(x) = f(y) = \sqrt{N}$. Then $f(x) = f(y) = f(x,z) = f(y,z) = \sqrt{N}$, $f(z) = 2\sqrt{N}$, and $f(x,y) = 0$. This yields $e_G(x,y) = \infty$ and $e_G(x,z) = e_G(z,y) = 2/\log N$, which violates the triangle inequality for all $N$. It follows that the NWD is not a metric.

Therefore, the liberation from lossless compression as in (6) to probabilities based on page counts as in (8) causes in certain cases the loss of metricity. But this is proper for a relative semantics. Indeed, we should view the distance $e_G$ between two concepts as a relative semantic similarity measure between those concepts. While concept $x$ is semantically close to concept $y$ and concept $y$ is semantically close to concept $z$, concept $x$ can be semantically very different from concept $z$.

Another important property of the NWD is its *scale-invariance* under the assumption that if the number $N$ of pages indexed by the search engine grows sufficiently large, the number of pages containing a given search term goes to a fixed fraction of $N$, and so does the number of pages containing conjunctions of search terms. This means that if $N$ doubles, then so do the $f$-frequencies. For the NWD to give us an objective semantic relation between search terms, it needs to become stable when the number $N$ of indexed pages grows. Some evidence that this actually happens was given in the example in Section 1.

The NWD can be used as a tool to investigate the meaning of terms and the relations between them as given by the Internet. This approach can be compared with the *Cyc* project [24], which tries to create artificial common sense. Cyc's knowledge base consists of hundreds of microtheories and hundreds of thousands of terms, as well as over a million hand-crafted assertions written in a formal language called CycL [33]. CycL is an enhanced variety of first order predicate logic. This knowledge base was created over the course of decades by paid human experts. It is therefore of extremely high quality. The Internet, on the other hand, is almost completely unstructured, and offers only a primitive query capability that is not nearly flexible enough to represent formal deduction. But what it lacks in expressiveness the Internet makes up for in size; Internet search engines have already indexed more than ten billion pages and show no signs of slowing down. Therefore search engine databases represent the largest publicly-available single corpus of aggregate statistical and indexing information so far created, and it seems that even rudimentary analysis thereof yields a variety of intriguing possibilities. It is unlikely, however, that this approach can ever achieve 100% accuracy like in principle deductive logic can, because the Internet mirrors humankind's own imperfect and varied nature. But, as we will see below, in practical terms the NWD can offer an easy way to provide results that are good enough for many applications, and which would be far too much work if not impossible to program in a deductive way.

In the following sections we present a number of applications of the NWD: hierarchical clustering and classification of concepts and names in a variety of domains, and finding corresponding words in different languages.

# 7 Applications and Experiments

To perform the experiments in this section, we used the *CompLearn* software tool [7]. The same tool has been used also to construct trees representing hierarchical clusters of objects in an unsupervised way using the normalized compression distance (NCD). However, now we use the normalized web distance (NWD).

## 7.1 Hierarchical Clustering

The method first calculates a distance matrix using the NWDs among all pairs of terms in the input list. Then it calculates a best-matching unrooted ternary tree using a novel quartet-method style heuristic based on randomized hill-climbing using a new fitness objective function optimizing the summed costs of all quartet topologies embedded in candidate trees [9]. Of course, given the distance matrix one can use also standard tree-reconstruction software from biological packages like the MOLPHY package [1].

However, such biological packages are based on data that are structured like rooted binary trees, and possibly do not perform well on hierarchical clustering of arbitrary natural data sets.

**Colors and numbers.** In the first example [10], the objects to be clustered are search terms consisting of the names of colors, numbers, and some words that are related but no color or number. The program automatically organized the colors towards one side of the tree and the numbers towards the other, Figure 1. It arranges the terms which have as only meaning a color or a number, and nothing else, on the farthest reach of the color side and the number side, respectively. It puts the more general terms black and white, and zero, one, and two, towards the center, thus indicating their more ambiguous interpretation. Also, things which were not exactly colors or numbers are also put towards the center, like the word "small." We may consider this an (admittedly very weak) example of automatic ontology creation.

**English novelists.** The authors and texts used are:

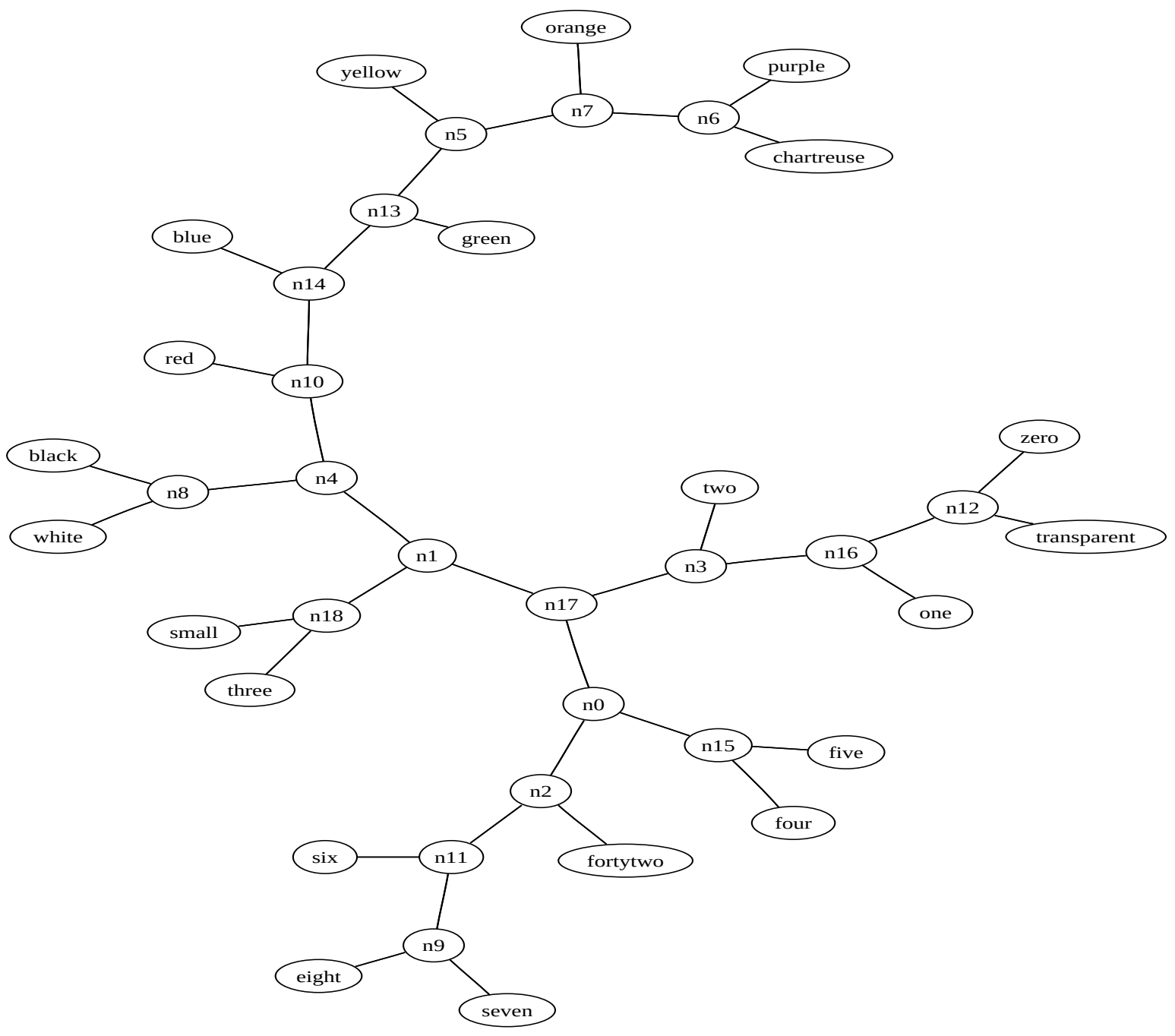

Figure 1: Colors, numbers, and other terms arranged into a tree based on the normalized web distances between the terms

William Shakespeare: *A Midsummer Night's Dream; Julius Caesar; Love's Labours Lost; Romeo and Juliet* .

Jonathan Swift: *The Battle of the Books; Gulliver's Travels; Tale of a Tub; A Modest Proposal*;

Oscar Wilde: *Lady Windermere's Fan; A Woman of No Importance; Salome; The Picture of Dorian Gray.*

The clustering is given in Figure 2, and to provide a feeling for the figures involved we give the associated NWD matrix in Figure 3. The $S(T)$ value written in Figure 2 gives the fidelity of the tree as a representation of the pairwise distances in the NWD matrix: $S(T) = 1$ is perfect and $S(T) = 0$ is as bad as possible. For details see [7, 9]

The question arises why we should expect this outcome. Are names of artistic objects so distinct? Yes. The point also being that the distances from every single object to all other objects are involved. The tree takes this global aspect into account and therefore disambiguates other meanings of the objects to retain the meaning that is relevant for this collection.

Is the distinguishing feature subject matter or title style? In these experiments with objects belonging to the cultural heritage it is clearly a subject matter. To stress the point we used "Julius Caesar" of Shakespeare. This term occurs on the web overwhelmingly in other contexts and styles. Yet the collection of the other objects used, and the semantic distance towards those objects, given by the NWD formula, singled out the semantics of "Julius Caesar" relevant to this experiment. Term co-occurrence in this specific context of author discussion is not swamped by other uses of this common term because of the particular form of the NWD and the distances being pairwise. Using very common book titles this swamping effect may still arise though.

Does the system gets confused if we add more artists? Representing the NWD matrix in bifurcating trees without distortion becomes more difficult for, say, more than 25 objects. See [9].

What about other subjects, like music or sculpture? Presumably, the system will be more trustworthy if the subjects are more common on the web.

These experiments are representative for those we have performed with the current software. We did not cherry pick the best outcomes. For example, all experiments with these three English writers, with different selections of four works of each, always yielded a tree so that we could draw a convex hull around the works of each author, without overlap.

The NWD method works independently of the alphabet, and even takes Chinese characters. In

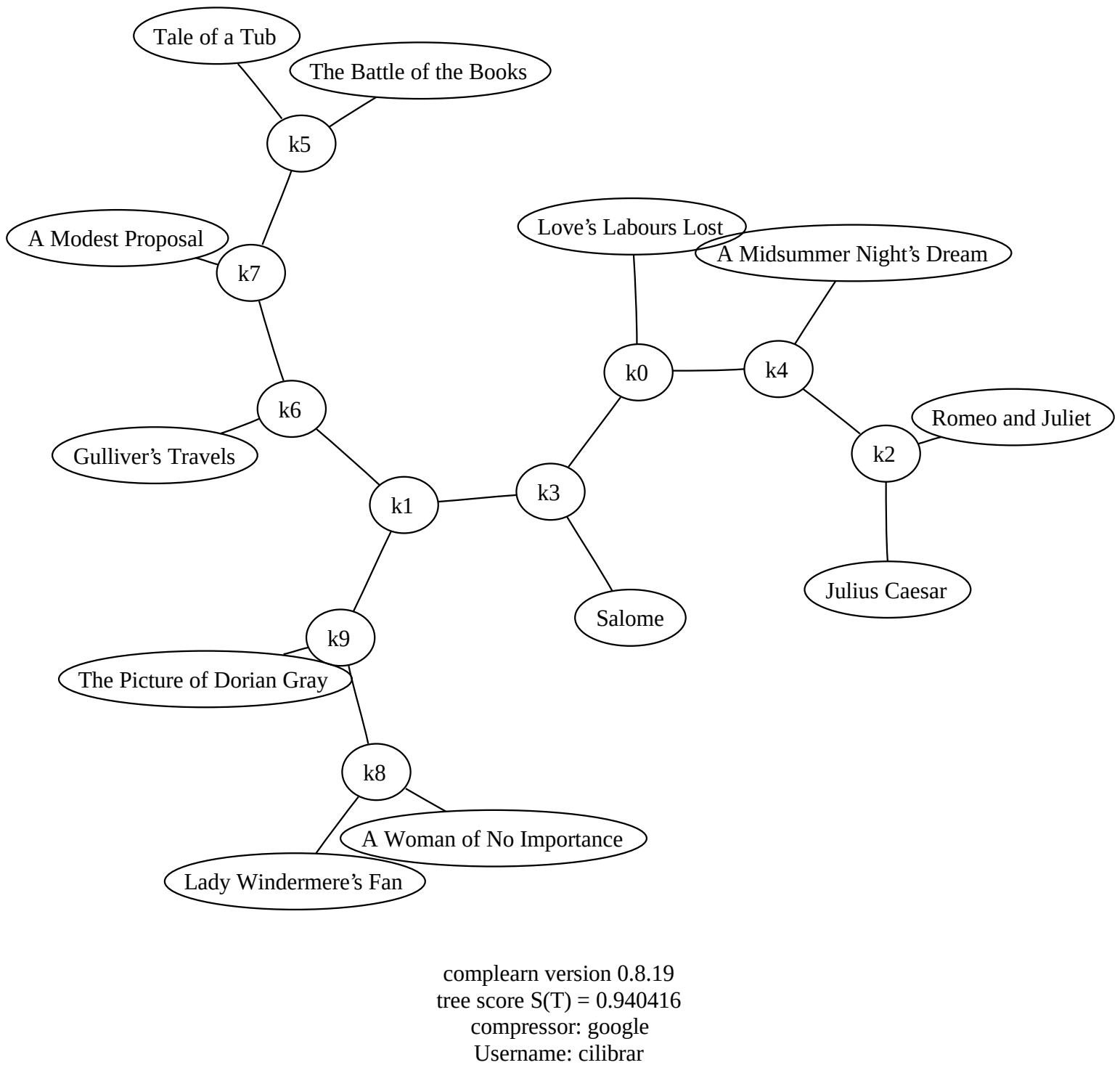

Figure 2: Hierarchical clustering of authors

| | | | | | | | | | | | | |
|---|---|---|---|---|---|---|---|---|---|---|---|---|
| A Woman of No Importance | 0.000 | 0.458 | 0.479 | 0.444 | 0.494 | 0.149 | 0.362 | 0.471 | 0.371 | 0.300 | 0.278 | 0.261 |
| A Midsummer Night's Dream | 0.458 | -0.011 | 0.563 | 0.382 | 0.301 | 0.506 | 0.340 | 0.244 | 0.499 | 0.537 | 0.535 | 0.425 |
| A Modest Proposal | 0.479 | 0.573 | 0.002 | 0.323 | 0.506 | 0.575 | 0.607 | 0.502 | 0.605 | 0.335 | 0.360 | 0.463 |
| Gulliver's Travels | 0.445 | 0.392 | 0.323 | 0.000 | 0.368 | 0.509 | 0.485 | 0.339 | 0.535 | 0.285 | 0.330 | 0.228 |
| Julius Caesar | 0.494 | 0.299 | 0.507 | 0.368 | 0.000 | 0.611 | 0.313 | 0.211 | 0.373 | 0.491 | 0.535 | 0.447 |
| Lady Windermere's Fan | 0.149 | 0.506 | 0.575 | 0.565 | 0.612 | 0.000 | 0.524 | 0.604 | 0.571 | 0.347 | 0.347 | 0.461 |
| Love's Labours Lost | 0.363 | 0.332 | 0.607 | 0.486 | 0.313 | 0.525 | 0.000 | 0.351 | 0.549 | 0.514 | 0.462 | 0.513 |
| Romeo and Juliet | 0.471 | 0.248 | 0.502 | 0.339 | 0.210 | 0.604 | 0.351 | 0.000 | 0.389 | 0.527 | 0.544 | 0.380 |
| Salome | 0.371 | 0.499 | 0.605 | 0.540 | 0.373 | 0.568 | 0.553 | 0.389 | 0.000 | 0.520 | 0.538 | 0.407 |
| Tale of a Tub | 0.300 | 0.537 | 0.335 | 0.284 | 0.492 | 0.347 | 0.514 | 0.527 | 0.524 | 0.000 | 0.160 | 0.421 |
| The Battle of the Books | 0.278 | 0.535 | 0.359 | 0.330 | 0.533 | 0.347 | 0.462 | 0.544 | 0.541 | 0.160 | 0.000 | 0.373 |
| The Picture of Dorian Gray | 0.261 | 0.415 | 0.463 | 0.229 | 0.447 | 0.324 | 0.513 | 0.380 | 0.402 | 0.420 | 0.373 | 0.000 |

Figure 3: Distance matrix of pairwise NWD's

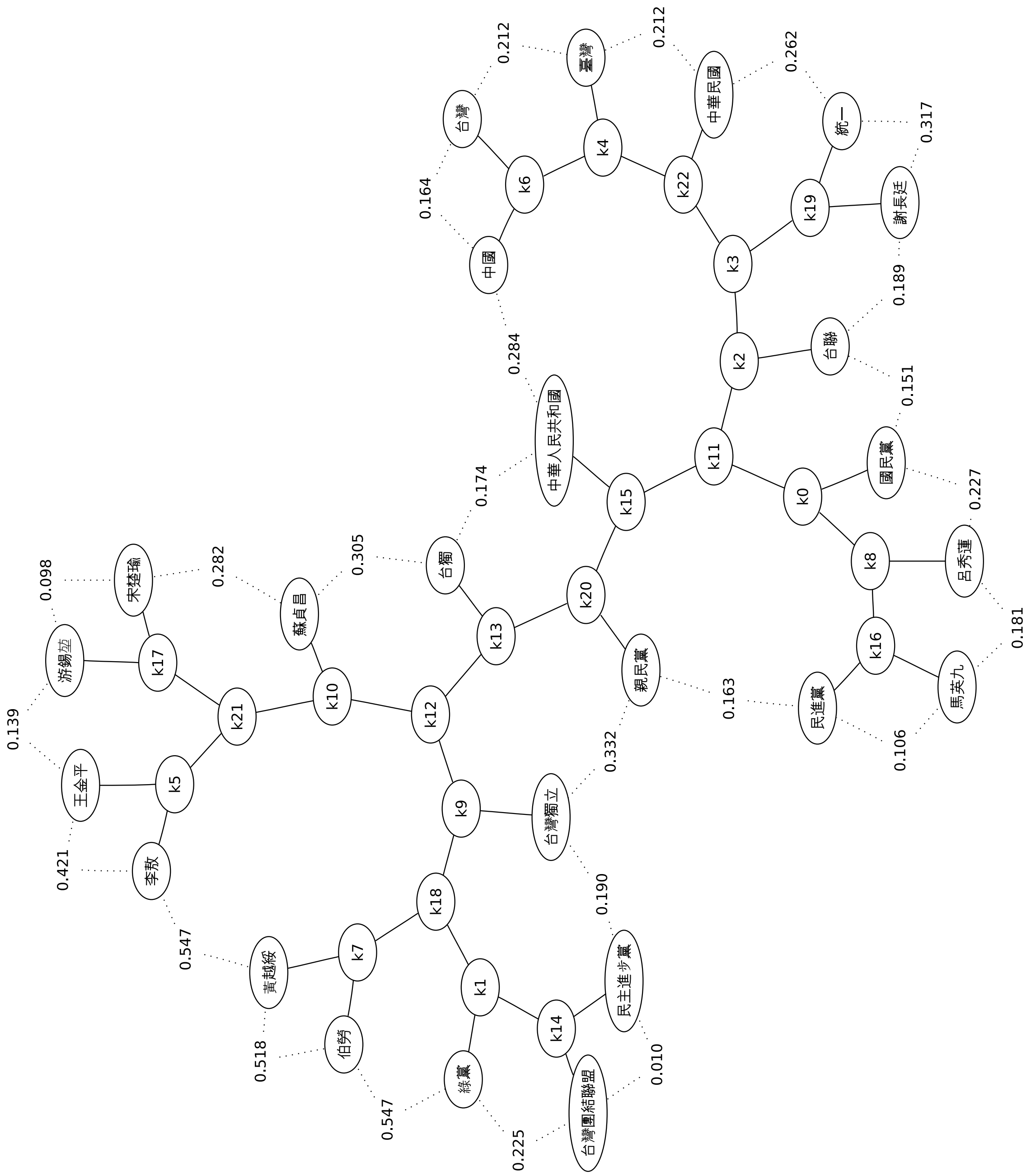

Figure 4: Names of several Chinese people, political parties, regions, and others. The nodes and *solid lines* constitute a tree constructed by a hierarchical clustering method based on the normalized web distances between all names. The numbers at the perimeter of the tree represent NWD values between the nodes pointed to by the *dotted lines*. For an explanation of the names, refer to Figure 5

中國 China
中華人民共和國 People's Republic of China
中華民國 Republic of China
伯勞 shrike (bird) [outgroup]
台灣 Taiwan (with simplified character "tai")
台灣團結聯盟 Taiwan Solidarity Union [Taiwanese political party]
台灣獨立 Taiwan independence
台獨 (abbreviation of the above)
台聯 (abbreviation of Taiwan Solidarity Union)
呂秀蓮 Annette Lu
國民黨 Kuomintang
宋楚瑜 James Soong
李敖 Li Ao
民主進步黨 Democratic Progressive Party
民進黨 (abbreviation of the above)
游錫堃 Yu Shyi-kun
王金平 Wang Jin-pyng
統一 unification [Chinese unification]
綠黨 Green Party
臺灣 Taiwan (with traditional character "tai")
蘇貞昌 Su Tseng-chang
親民黨 People First Party [political party in Taiwan]
謝長廷 Frank Hsieh
馬英九 Ma Ying-jeou
黃越綏 a presidential advisor and 2008 presidential hopeful

Figure 5: Explanations of the Chinese names used in the experiment that produced Figure 4, courtesy Dr. Kaihsu Tai

## Training Data

| | | | | |
|---|---|---|---|---|
| *Positive Training* | (22 cases) | | | |
| avalanche | bomb threat | broken leg | burglary | car collision |
| death threat | fire | flood | gas leak | heart attack |
| hurricane | landslide | murder | overdose | pneumonia |
| rape | roof collapse | sinking ship | stroke | tornado |
| train wreck | trapped miners | | | |
| *Negative Training* | (25 cases) | | | |
| arthritis | broken dishwasher | broken toe | cat in tree | contempt of court |
| dandruff | delayed train | dizziness | drunkenness | enumeration |
| flat tire | frog | headache | leaky faucet | littering |
| missing dog | paper cut | practical joke | rain | roof leak |
| sore throat | sunset | truancy | vagrancy | vulgarity |
| *Anchors* | (6 dimensions) | | | |
| crime | happy | help | safe | urgent |
| wash | | | | |

## Testing Results

| | Positive tests | Negative tests |
|---|---|---|
| Positive Predictions | assault, coma, electrocution, heat stroke, homicide, looting, meningitis, robbery, suicide | menopause, prank call, pregnancy, traffic jam |
| Negative Predictions | sprained ankle | acne, annoying sister, campfire, desk, mayday, meal |

## Accuracy

15/20 = 75.00%

Figure 6: NWD–SVM learning of "emergencies."

the example of Figure 4, several Chinese names were entered. The tree shows the separation according to concepts like regions, political parties, people, etc. See Figure 5 for English translations of these names. The dotted lines with numbers between each adjacent node along the perimeter of the tree represent the NWD values between adjacent nodes in the final ordered tree. The tree is presented in such a way that the sum of these values in the entire ring is minimized. This generally results in trees that make the most sense upon initial visual inspection, converting an unordered bifurcating tree to an ordered one. This feature allows for a quick visual inspection around the edges to determine the major groupings and divisions among coarse structured problems.

## 7.2 Classification

In cases in which the set of objects can be large, in the millions, clustering cannot do us much good. We may also want to do definite classification, rather than the more fuzzy clustering. To this purpose, we augment the search engine method by adding a trainable component of the learning system. Here we use the Support Vector Machine (SVM) as a trainable component. For the SVM method used in this paper, we refer to the survey [5]. One can use the $e_G$ distances as an oblivious feature-extraction technique to convert generic objects into finite-dimensional vectors.

Let us consider a binary classification problem on examples represented by search terms. In these experiments we require a human expert to provide a list of, say, 40 *training words*, consisting of half positive examples and half negative examples, to illustrate the contemplated concept class. The expert also provides, say, six *anchor words* $a_1, \ldots, a_6$, of which half are in some way related to the concept under consideration. Then, we use the anchor words to convert each of the 40 training words $w_1, \ldots, w_{40}$ to 6-dimensional *training vectors* $\bar{v}_1, \ldots, \bar{v}_{40}$. The entry $v_{j,i}$ of $\bar{v}_j = (v_{j,1}, \ldots, v_{j,6})$ is defined as $v_{j,i} = e_G(w_j, a_i)$ $(1 \leq j \leq 40, 1 \leq i \leq 6)$. The training vectors are then used to train an SVM to learn the concept, and then test words may be classified using the same anchors and trained SVM model. Finally testing is performed using 20 examples in a balanced ensemble to yield a final accuracy. The kernel-width and error-cost parameters are automatically determined using five-fold cross validation. The LIBSVM software [6] was used for all SVM experiments.

**Classification of "emergencies."** In Figure 6, we trained using a list of "emergencies" as positive examples, and a list of "almost emergencies" as negative examples. The figure is self-explanatory. The accuracy on the test set is 75%.

**Classification of prime numbers.** In an experiment to learn prime numbers, we used the literal search terms below (digital numbers and alphabetical words) in the Google search engine.
*Positive training examples*: 11, 13, 17, 19, 2, 23, 29, 3, 31, 37, 41, 43, 47, 5, 53, 59, 61, 67, 7, 71, 73.
*Negative training examples*: 10, 12, 14, 15, 16, 18, 20, 21, 22, 24, 25, 26, 27, 28, 30, 32, 33, 34, 4, 6, 8, 9.

*Anchor words*: composite, number, orange, prime, record.
*Unseen test examples*: The numbers 101, 103, 107, 109, 79, 83, 89, 97 were correctly classified as primes. The numbers 36, 38, 40, 42, 44, 45, 46, 48, 49 were correctly classified as nonprimes. The numbers 91 and 110 were false positives, since they were incorrectly classified as a primes. There were no false negatives. The accuracy on the test set is 17/19 = 89.47%. Thus, the method learns to distinguish prime numbers from nonprime numbers by example, using a search engine. This example illustrates several common features of our method that distinguish it from the strictly deductive techniques.

## 7.3 Matching the Meaning

Assume that there are five words that appear in two different matched sentences, but the permutation associating the English and Spanish words is, as yet, undetermined. Let us say, *plant, car, dance, speak, friend* versus *bailar, hablar, amigo, coche, planta*. At the outset we assume a preexisting vocabulary of eight English words with their matched Spanish translations: *tooth, diente; joy, alegria; tree, arbol; electricity, electricidad; table, tabla; money, dinero; sound, sonido; music, musica*. Can we infer the correct permutation mapping the unknown words using the preexisting vocabulary as a basis?

We start by forming an English basis matrix in which each entry is the $e_G$ distance between the English word labeling the column and the English word labeling the row. We label the columns by the translation-known English words, and the rows by the translation-unknown English words. Next, we form a Spanish matrix with the known Spanish words labeling the columns in the same order as the known English words. But now we label the rows by choosing one of the many possible permutations of the unknown Spanish words. For every permutation, each matrix entry is the $e_G$ distance between the Spanish words labeling the column and the row. Finally, choose the permutation with the highest positive correlation between the English basis matrix and the Spanish matrix associated with the permutation. If there is no positive correlation report a failure to extend the vocabulary. The method inferred the correct permutation for the testing words: *plant, planta; car, coche; dance, bailar; speak, hablar; friend, amigo.*

## 7.4 Systematic Comparison with WordNet Semantics

WordNet [17] is a semantic concordance of English. It focuses on the meaning of words by dividing them into categories. We use this as follows. A category we want to learn, the concept, is termed, say, "electrical", and represents anything that may pertain to electrical devices. The negative examples are constituted by simply everything else. This category represents a typical expansion of a node in the WordNet hierarchy. In an experiment we ran, the accuracy on this test set is 100%: It turns out that "electrical terms" are unambiguous and easy to learn and classify by our method.

The information in the WordNet database is entered over the decades by human experts and is precise. The database is an academic venture and is publicly accessible. Hence it is a good baseline against which to judge the accuracy of our method in an indirect manner. While we cannot directly compare the semantic distance, the NWD, between objects, we can indirectly judge how accurate it is by using it as basis for a learning algorithm. In particular, we investigated how well semantic categories as learned using the NWD–SVM approach agree with the corresponding WordNet categories. For details about the structure of WordNet we refer to the official WordNet documentation available online.

We considered 100 randomly selected semantic categories from the WordNet database. For each category we executed the following sequence. First, the SVM is trained on 50 labeled training samples. The positive examples are randomly drawn from the WordNet database in the category in question. The negative examples are randomly drawn from a dictionary. While the latter examples may be false negatives, we consider the probability negligible. Per experiment we used a total of six anchors, three of which are randomly drawn from the WordNet database category in question, and three of which are drawn from the dictionary. Subsequently, every example is converted to 6-dimensional vectors using NWD. The $i$th entry of the vector is the NWD between the $i$th anchor and the example concerned ($1 \leq i \leq 6$). The SVM is trained on the resulting labeled vectors. The kernel-width and error-cost parameters are automatically determined using five-fold cross validation. Finally, testing of how well the SVM has learned the classifier is performed using 20 new examples in a balanced ensemble of positive and negative examples obtained in the same way, and converted to 6-dimensional vectors in the same manner, as the training examples. This results in an accuracy score of correctly classified test examples. We ran 100 experiments. The actual data are available at [8].

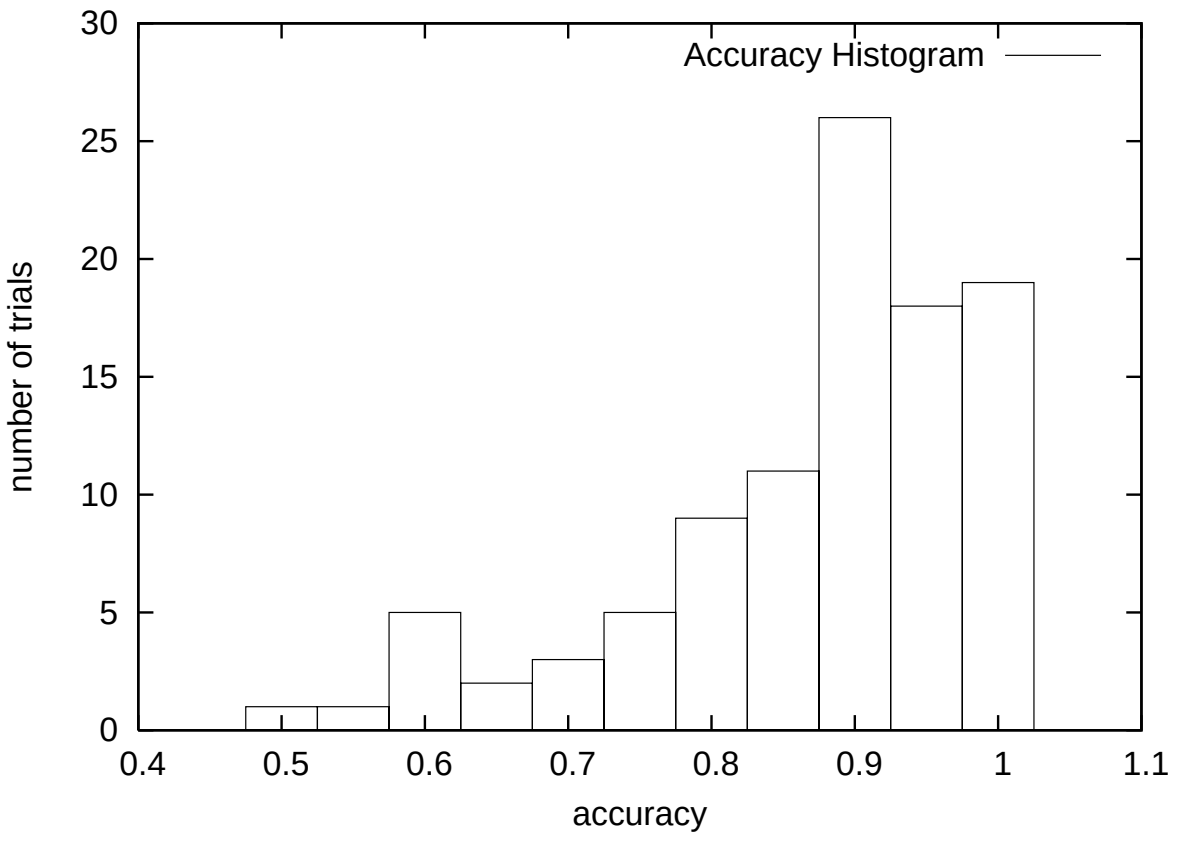

Figure 7: Histogram of accuracies over 100 trials of WordNet experiment.

A histogram of agreement accuracies is shown in Figure 7. On average, our method turns out to agree well with the WordNet semantic concordance made by human experts. The mean of the accuracies of agreements is 0.8725. The variance is $\approx 0.01367$, which gives a standard deviation of $\approx 0.1169$. Thus, it is rare to find agreement less than 75%. The total number of web searches involved in this randomized automatic trial is upper bounded by $100 \times 70 \times 6 \times 3 = 126,000$. A considerable savings resulted from the fact that it is simple to cache search count results for efficiency. For every new term, in computing its 6-dimensional vector, the NWD computed with respect to the six anchors requires the counts for the anchors which needs to be computed only once for each experiment, the count of the new term which can be computed once, and the count of the joint occurrence of the new term and each of the six anchors, which has to be computed in each case. Altogether, this gives a total of $6 + 70 + 70 \times 6 = 496$ for every experiment, so $49,600$ web searches for the entire trial.

# 8 Conclusion

The approach in this chapter rests on the idea that *information distance* between two objects can be measured by the size of the shortest description that transforms each object into the other one. This idea is most naturally expressed mathematically using Kolmogorov complexity. Kolmogorov complexity, moreover, provides mathematical tools to show that such a measure is, in a proper sense, universal among all (upper semi)computable distance measures satisfying a natural density condition. These comprise most, if not all, distances one may be interested in. Since two large, very similar, objects may have the same information distance as two small, very dissimilar, objects, in terms of similarity it is the relative distance we are interested in. Hence we normalize the information metric to create a relative similarity in between 0 and 1. However, the normalized information metric is uncomputable. We approximate its Kolmogorov complexity parts by off the shelve compression programs (in the case of the normalized compression distance) or readily available statistics from the Internet (in case of the normalized web distance). The outcomes are two practical distance measures for literal as well as for non-literal data that have been proved useful in numerous applications, some of which have been presented in the previous sections.

It is interesting that while the (normalized) information distance and the normalized compression distance between literal objects are metrics, this is *not* the case for the normalized web distance or NWD between nonliteral objects like words, which is the measure of word similarity that we use here. The latter derives the code-word lengths involved from statistics gathered from the Internet or another large database with an associated search engine that returns aggregate page counts or something similar. This has two effects: (i) the code-word length involved is one that on average is shortest for the probability involved, and (ii) the statistics involved are related to hits on Internet pages and not to genuine probabilities. For example, if every page containing term $x$ also contains term $y$ and vice versa, then the NWD between $x$ and $y$ is 0, even though $x$ and $y$ may be different (like "yes" and "no"). The consequence is that the NWD distance takes values primarily (but not exclusively) in $[0, 1]$ and is not a metric. Thus, while 'name1' is semantically close to 'name2,' and 'name2' is semantically close to 'name3,' 'name1' can be semantically very different from 'name3.' This is as it should be for a relative semantics: while 'man' is close to 'centaur', and 'centaur' is close to 'horse,' 'man' is far removed from 'horse' [40].

The NWD can be compared with the *Cyc* project [24] or the WordNet project [17]. These projects try to create artificial common sense. The knowledge bases involved were created over the course of decades by paid human experts. They are therefore of extremely high quality. An aggregate page count returned by a search engine, on the other hand, is almost completely unstructured, and offers only a primitive query capability that is not nearly flexible enough to represent formal deduction. But what it lacks in expressiveness a search engine makes up for in size; many search engines already index more than ten billion pages and more data comes online every day.

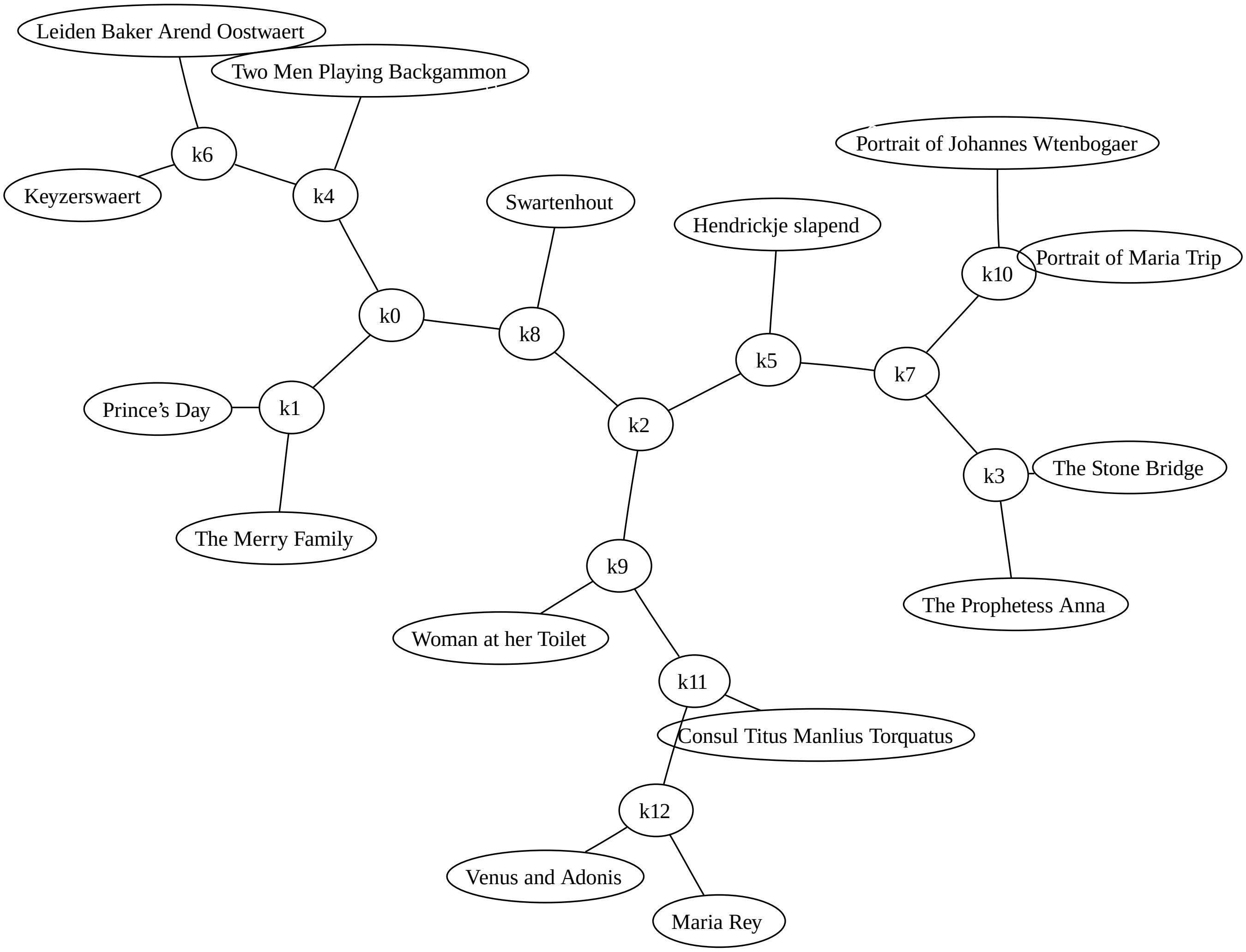

complearn version 0.8.19
tree score S(T) = 0.940019
compressor: google
Username: cilibrar